\documentclass[11pt,a4paper]{article}
\usepackage[hyperref]{emnlp2020}
\usepackage{times}
\usepackage{latexsym}
\usepackage{url}
\usepackage{graphicx}
\usepackage{amsmath}
\usepackage{amssymb}
\usepackage{subfigure}
\usepackage{caption}
\usepackage{algorithm}
\usepackage{algorithmicx}
\usepackage{algpseudocode}
\usepackage{multicol}
\usepackage{multirow}
\usepackage{booktabs}
\usepackage{listings}
\usepackage{bbm}
\usepackage{booktabs}
\usepackage{array}
\usepackage{bm}

\newcommand{\tworow}[1]{\multirow{2}{*}{#1}}
\newcommand{\twocol}[1]{\multicolumn{2}{c}{#1}}

\usepackage{microtype}

\aclfinalcopy 

\title{Dynamic Anticipation and Completion for \\ Multi-Hop Reasoning over Sparse Knowledge Graph}

  \newcommand*{\email}[1]{\texttt{#1}}

  \author{
    \textbf{Xin Lv}$^{1,2}$, \textbf{Xu Han}$^{1,2}$, \textbf{Lei Hou}$^{1,2}$\thanks{\quad Corresponding Author}\hspace{0.5em}, \textbf{Juanzi Li}$^{1,2}$, \textbf{Zhiyuan Liu}$^{1,2}$ \\ \textbf{Wei Zhang}$^{3}$, \textbf{Yichi Zhang}$^{3}$, \textbf{Hao Kong}$^{3}$, \textbf{Suhui Wu}$^{3}$\\
    $^1$Department of Computer Science and Technology, BNRist \\
    $^2$KIRC, Institute for Artificial Intelligence, Tsinghua University, Beijing 100084, China \\
     $^3$Alibaba Group, Hangzhou, China \\
    \email{\{lv-x18,hanxu17\}@mails.tsinghua.edu.cn}\\
    \email{\{houlei,lijuanzi,liuzy\}@tsinghua.edu.cn}
    }

\date{}

\begin{document}
\maketitle
\begin{abstract}
Multi-hop reasoning has been widely studied in recent years to seek an effective and interpretable method for knowledge graph (KG) completion. 
Most previous reasoning methods are designed for dense KGs with enough paths between entities, but cannot work well on those sparse KGs that only contain sparse paths for reasoning. 
On the one hand, sparse KGs contain less information, which makes it difficult for the model to choose correct paths. On the other hand, the lack of evidential paths to target entities also makes the reasoning process difficult.
To solve these problems, we propose a multi-hop reasoning model named \textbf{DacKGR} over sparse KGs, by applying novel dynamic anticipation and completion strategies: (1) The anticipation strategy utilizes the latent prediction of embedding-based models to make our model perform more potential path search over sparse KGs. (2) Based on the anticipation information, the completion strategy dynamically adds edges as additional actions during the path search, which further alleviates the sparseness problem of KGs. The experimental results on five datasets sampled from Freebase, NELL and Wikidata show that our method outperforms state-of-the-art baselines. Our codes and datasets can be obtained from \url{https://github.com/THU-KEG/DacKGR}.

\end{abstract}

\section{Introduction}

Knowledge graphs (KGs) represent the world knowledge in a structured way, and have been proven to be helpful for many downstream NLP tasks like query answering \cite{traversing}, dialogue generation \cite{dialogue} and machine reading comprehension \cite{reading}. Despite their wide applications, many KGs still face serious incompleteness \cite{TransE}, which limits their further development and adaption for related downstream tasks. 

\begin{figure}[t]
  \centering
  \setlength{\abovecaptionskip}{2pt}
  \setlength{\belowcaptionskip}{0pt}
  \includegraphics[width=75mm]{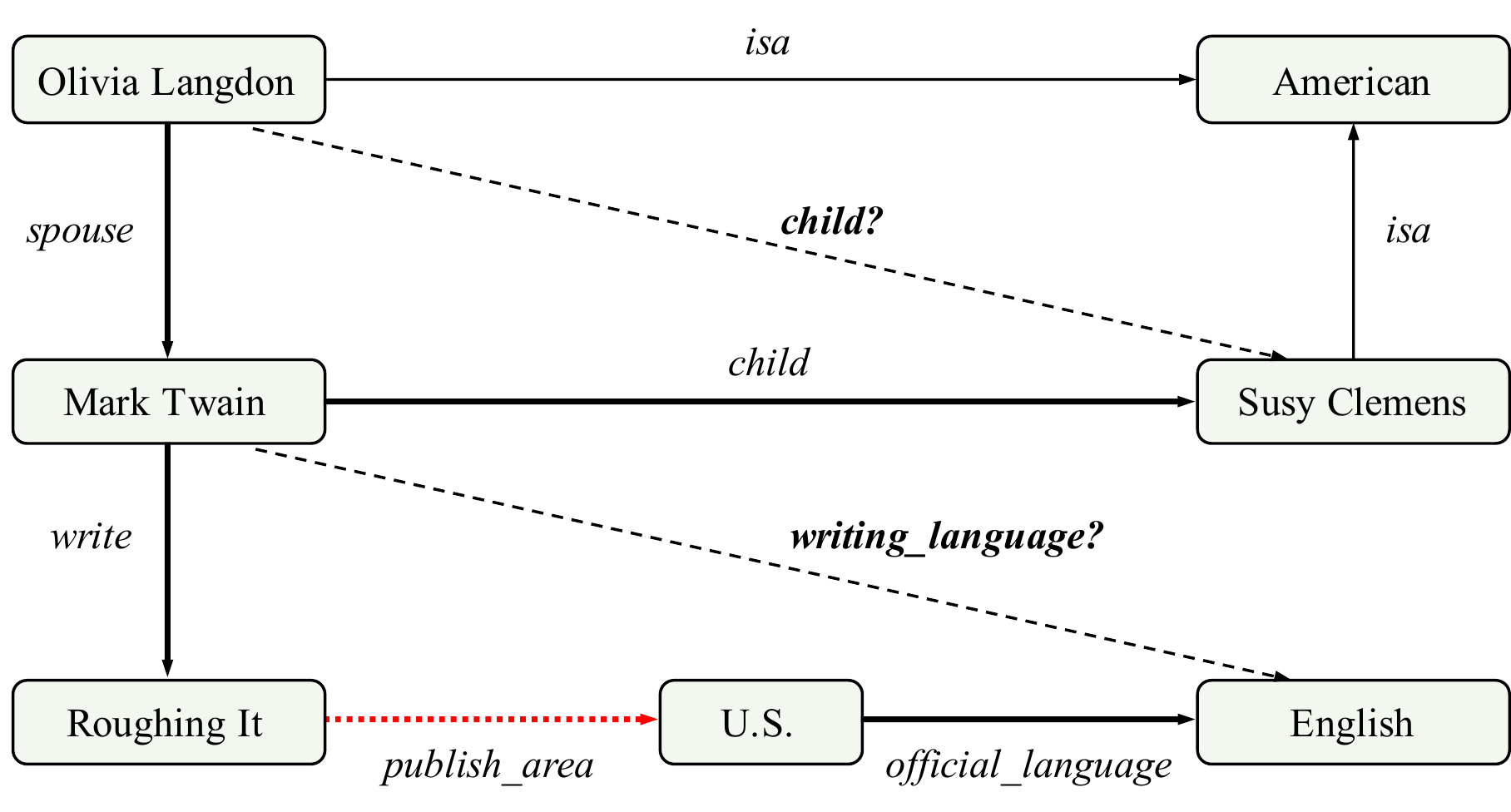}
  \caption{An illustration of multi-hop reasoning task over sparse KG. The missing relations (black dashed arrows) between entities can be inferred from existing triples (solid black arrows) through reasoning paths (bold arrows). However, some relations in the reasoning path are missing (red dashed arrows) in sparse KG, which makes multi-hop reasoning difficult. }
  \label{multihop_reasoning}
  \end{figure}

To alleviate this issue, some embedding-based models \cite{TransE, ConvE} are proposed, most of which embed entities and relations into a vector space and make link predictions to complete KGs. These models focus on efficiently predicting knowledge but lack necessary interpretability. 
In order to solve this problem, \citet{MINERVA} and \citet{MultiHop} propose multi-hop reasoning models, which use the REINFORCE algorithm \cite{RL} to train an agent to search over KGs. These models can not only give the predicted result but also an interpretable path to indicate the reasoning process. 
As shown in the upper part of Figure \ref{multihop_reasoning}, for a triple query (\textit{Olivia Langdon}, \textit{child}, ?), multi-hop reasoning models can predict the tail entity \textit{Susy Clemens} through a reasoning path (bold arrows).

Although existing multi-hop reasoning models have achieved good results, they still suffer two problems on sparse KGs: (1) \textbf{Insufficient information}. Compared with normal KGs, sparse KGs contain less information, which makes it difficult for the agent to choose the correct search direction. (2) \textbf{Missing paths}. In sparse KGs, some entity pairs do not have enough paths between them as reasoning evidence, which makes it difficult for the agent to carry out the reasoning process.
As shown in the lower part of Figure \ref{multihop_reasoning}, there is no evidential path between \textit{Mark Twain} and \textit{English} since the relation \textit{publish\_area} is missing. 
From Table \ref{dataset_statistic} we can learn that some sampled KG datasets are actually sparse.
Besides, some domain-specific KGs (e.g., WD-singer) do not have abundant knowledge and also face the problem of sparsity.

As the performance of most existing multi-hop reasoning methods drops significantly on sparse KGs, some preliminary efforts, such as CPL \cite{CPL}, explore to introduce additional text information to ease the sparsity of KGs. Although these explorations have achieved promising results, they are still limited to those specific KGs whose entities have additional text information. Thus, reasoning over sparse KGs is still an important but not fully resolved problem, and requires a more generalized approach to this problem.

\begin{table}
  \small
  \centering
  \setlength\tabcolsep{4pt}
  \begin{tabular}{lrrrrr}
    \toprule
    \tworow{\textbf{Dataset}} & \tworow{\textbf{\#Ent}} & \tworow{\textbf{\#Rel}} & \tworow{\textbf{\#Fact}} & \twocol{\textbf{\#degree}} \\
    \cline{5-6}
      & & & & \textbf{mean} & \textbf{median} \\
      \midrule 
      FB15K-237 & 14,505 & 237 & 272,115 & 19.74 & 14 \\
      WN18RR & 40,945 & 11 & 86,835 & 2.19 & 2 \\
      NELL23K & 22,925 & 200 & 35,358 & 2.21 & 1 \\
      WD-singer & 10,282 & 135 & 20,508 & 2.35 & 2 \\
      \bottomrule
  \end{tabular}
  \caption{The statistics of some benchmark KG datasets. \#degree is the outgoing degree of every entity that can indicate the sparsity level.}
  \label{dataset_statistic}
\end{table}

In this paper, we propose a multi-hop reasoning model named DacKGR, along with two dynamic strategies to solve the two problems mentioned above:

\textbf{Dynamic Anticipation} makes use of the limited information in a sparse KG to anticipate potential targets before the reasoning process. 
Compared with multi-hop reasoning models, embedding-based models are robust to sparse KGs, because they depend on every single triple rather than paths in KG. 
To this end, our anticipation strategy injects the pre-trained embedding-based model's predictions as anticipation information into the states of reinforcement learning. This information can guide the agent to avoid aimlessly searching paths.

 \textbf{Dynamic Completion} temporarily expands the part of a KG to enrich the options of path expansion during the reasoning process. In sparse KGs, many entities only have few relations, which limits the choice spaces of the agent. Our completion strategy thus dynamically adds some additional relations (e.g., red dashed arrows in Figure \ref{multihop_reasoning}) according to the state information of the current entity during searching reasoning paths. After that, for the current entity and an additional relation $r$, we use a pre-trained embedding-based model to predict tail entity $e$. Then, the additional relation $r$ and the predicted tail entity $e$ will form a potential action $(r, e)$ and be added to the action space of the current entity for path expansion. 
 
We conduct experiments on five datasets sampled from Freebase, NELL and Wikidata. The results show that our model DacKGR outperforms previous multi-hop reasoning models, which verifies the effectiveness of our model.

\section{Problem Formulation}

In this section, we first introduce some symbols and concepts related to normal multi-hop reasoning, and then formally define the task of multi-hop reasoning over sparse KGs.

\textit{Knowledge graph} $\mathcal{KG}$ can be formulated as $\mathcal{KG} = \{\mathcal{E}, \mathcal{R}, \mathcal{T}\}$, where $\mathcal{E}$ and $\mathcal{R}$ denote entity set and relation set respectively. $\mathcal{T} = \{(e_s, r_q, e_o)\} \subseteq \mathcal{E} \times \mathcal{R} \times \mathcal{E}$ is triple set, where $e_s$ and $e_o$ are head and tail entities respectively, and $r_q$ is the relation between them.
For every KG, we can use the average out-degree $D_{avg}^{out}$ of each entity (node) to define its sparsity. Specifically, if $D_{avg}^{out}$ of a KG is larger than a threshold, we can say it is a dense or normal KG, otherwise, it is a sparse KG.

Given a graph $\mathcal{KG}$ and a triple query $(e_s, r_q, ?)$, where $e_s$ is the source entity and $r_q$ is the query relation, \textit{multi-hop reasoning for knowledge graphs} aims to predict the tail entity $e_o$ for $(e_s, r_q, ?)$. Different from previous KG embedding tasks, multi-hop reasoning also gives a supporting path $\{(e_s, r_1, e_1), (e_1, r_2, e_2)\dots, (e_{n-1}, r_n, e_o)\}$ over $\mathcal{KG}$ as evidence. As mentioned above, we mainly focus on the multi-hop reasoning task over sparse KGs in this paper.

\begin{figure*}[ht]
  \centering
  \setlength{\abovecaptionskip}{2pt}
  \setlength{\belowcaptionskip}{0pt}
  \includegraphics[width=160mm]{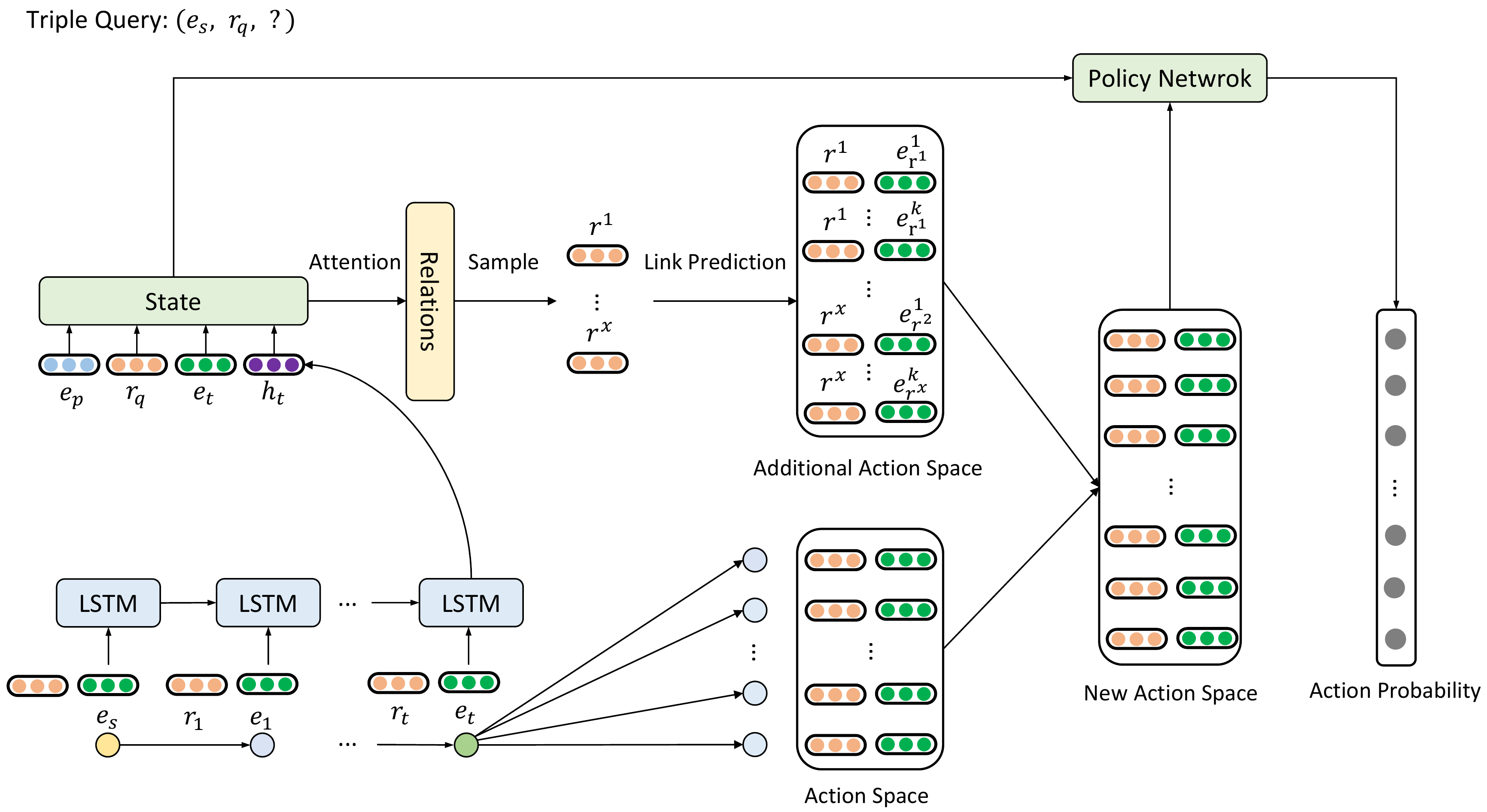} 
  \caption{An illustration of our policy network with dynamic anticipation and dynamic completion strategies. The vector of $e_p$ is the prediction information introduced in Section \ref{sec:state_prediction}. We use the current state to dynamically select some relations, and use the pre-trained embedding-based model to perform link prediction to obtain additional action space. The original action space will be merged with the additional action space to form a new action space.} 
  \label{pic:framework}
  \end{figure*}

\section{Methodology}

In this section, we first introduce the whole reinforcement learning framework for multi-hop reasoning, and then detail our two strategies designed for the sparse KGs, i.e., dynamic anticipation and dynamic completion.
The former strategy introduces the guidance information from embedding-based models to help multi-hop models find the correct direction on sparse KGs. Based on this strategy, the dynamic completion strategy introduces some additional actions during the reasoning process to increase the number of paths, which can alleviate the sparsity of KGs. Following \citet{MultiHop}, the overall framework of DacKGR is illustrated in Figure \ref{pic:framework}.

\subsection{Reinforcement Learning Framework}
\label{sec:framework}

In recent years, multi-hop reasoning for KGs has been formulated as a Markov Decision Process (MDP) over $\mathcal{KG}$ \cite{MINERVA}: given a triple query $(e_s, r_q, ?)$, the agent needs to start from the head entity $e_s$, continuously select the edge (relation) corresponding to the current entity with maximum probability as the direction, and jump to the next entity until the maximum number of hops $T$. Following previous work \cite{MultiHop}, the MDP consists of the following components:

\textbf{State} \indent In the process of multi-hop reasoning, which edge (relation) the agent chooses depends not only on the query relation $r_q$ and the current entity $e_t$, but also on the previous historical searching path. Therefore, the state of the $t$-th hop can be defined as $s_t = (r_q, e_t, h_t)$, where $h_t$ is the representation of the historical path. Specifically, we use an LSTM to encode the historical path information, $h_t$ is the output of LSTM at the $t$-th step.

\textbf{Action} \indent For a state $s_t = (r_q, e_t, h_t)$, if there is a triple $(e_t, r_n, e_n)$ in the KG, $(r_n, e_n)$ is an action of the state $s_t$. All actions of the state $s_t$ make up its action space $\mathcal{A}_t = \{(r, e) | (e_t, r, e) \in \mathcal{T}\}$. Besides, for every state $s_t$, we also add an additional action $(r_{\textit{LOOP}}, e_t)$, where $\textit{LOOP}$ is a manually added self-loop relation. It allows the agent to stay at the current entity, which is similar to a ``STOP'' action.

\textbf{Transition} \indent If the current state is $s_t = (r_q, e_t, h_t)$ and the agent chooses $(r_n, e_n) \in \mathcal{A}_t$ as the next action, then the current state $s_t$ will be converted to $s_{t+1} = (r_q, e_n, h_{t+1})$. In this paper, we limit the maximum number of hops to $T$, and the transition will end at the state $s_T = (r_q, e_T, h_T)$.

\textbf{Reward} \indent For a given query $(e_s, r_q, ?)$ with the golden tail entity $e_o$, if the agent finally stops at the correct entity, i.e., $e_T = e_o$, the reward is one, otherwise, the reward is a value between 0 and 1 given by the function $f(e_s, r_q, e_T)$, where the function $f$ is given by a pre-trained knowledge graph embedding (KGE) model for evaluating the correctness of the triple $(e_s, r_q, e_T)$.

\subsection{Policy Network}

For the above MDP, we need a policy network to guide the agent to choose the correct action in different states. 

We represent entities and relations in $\mathcal{KG}$ as vectors in a semantic space, and then the action $(r, e)$ at the step $t$ can be represented as $\mathbf{a}_t = [\mathbf{r}; \mathbf{e}]$, where $\mathbf{r}$ and $\mathbf{e}$ are the vectors of $r$ and $e$ respectively. As we mentioned in Section \ref{sec:framework}, we use an LSTM to store the historical path information. Specifically, the representation of each action selected by the agent will be fed into the LSTM to generate historical path information so far,
\begin{equation}
  \mathbf{h}_t = \text{LSTM}(\mathbf{h}_{t - 1}, \mathbf{a}_{t - 1}) .
\end{equation}
The representation of the $t$-th state $s_t = (r_q, e_t, h_t)$ can be formulated as 
\begin{equation}
\label{eq:state}
  \mathbf{s}_t = [\mathbf{r}_q; \mathbf{e}_t;  \mathbf{h}_t] .
\end{equation}
After that, we represent the action space by stacking all actions in $\mathcal{A}_t$ as $\mathbf{A}_t \in \mathbb{R}^{|\mathcal{A}_t| \times 2d}$, where $d$ is the dimension of the entity and relation vector. The policy network is defined as,
\begin{equation}
  \pi_{\theta} (a_t | s_t) = \sigma(\mathbf{A}_t(\mathbf{W}_1 \text{ReLU} (\mathbf{W}_2  \mathbf{s}_t))) ,
\end{equation}
where $\sigma$ is the softmax operator, $\mathbf{W}_1$ and $\mathbf{W}_2$ are two linear neural networks, and $\pi_{\theta} (a_t | s_t)$ is the probability distribution over all actions in $\mathcal{A}_t$.

\subsection{Dynamic Anticipation}
\label{sec:state_prediction}

As reported in previous work \cite{MINERVA,MultiHop}, although the KGE models are not interpretable, they can achieve better results than the multi-hop reasoning models on most KGs. This phenomenon is more obvious on the sparse KG (refer to experimental results in Table \ref{fb15k-lp}) since KGE models are more robust as they do not rely on the connectivity of the KGs.

Inspired by the above phenomenon, we propose a new strategy named \textit{dynamic anticipation}, which introduces the prediction information of the embedding-based models into the multi-hop reasoning models to guide the model learning. Specifically, for a triple query $(e_s, r_q, ?)$, we use the pre-trained KGE models to get the probability vector of all entities being the tail entity. Formally, the probability vector can be formulated as $\mathbf{p} \in \mathbb{R}^{|\mathcal{E}|}$, where the value of the $i$-th dimension of $\mathbf{p}$ represents the probability that $e_i$ is the correct tail entity.

For the dynamic anticipation strategy, we change the state representation in Equation \ref{eq:state} to:
\begin{equation}
  \mathbf{s}_t = [\mathbf{e}_p; \mathbf{r}_q; \mathbf{e}_t; \mathbf{h}_t] ,
\end{equation}
where $\mathbf{e}_p$ is prediction information given by KGE models. In this paper, we use the following three strategies to generate $\mathbf{e}_p$: (1) Sample strategy. We sample an entity based on probability distribution $\mathbf{p}$ and denote its vector as $\mathbf{e}_p$. (2) Top-one strategy. We select the entity with the highest probability in $\mathbf{p}$. (3) Average strategy. We take the weighted average of the vectors of all entities according to the probability distribution $\mathbf{p}$ as the prediction information $\mathbf{e}_p$. 
In experiments, we choose the strategy that performs best on the valid set.

\subsection{Dynamic Completion}
\label{sec:action_selection}

In sparse KGs, there are often insufficient evidential paths between head and tail entities, so that the performance of multi-hop reasoning models will drop significantly.

In order to solve the above problems, we propose a strategy named \textit{dynamic completion} to dynamically augment the action space of each entity during reasoning process. Specifically, for the current state $s_t$, its candidate set of additional actions can be defined as $C_{t} = \{(r, e) | r \in \mathcal{R} \wedge e \in \mathcal{E} \wedge (e_t, r, e) \not\in \mathcal{T}\}$. We need to select some actions with the highest probability from $C_{t}$ as additional actions, where the probability can be defined as:
\begin{equation}
  p((r, e) | s_t) = p(r | s_t) p(e | r, s_t) .
\end{equation}
However, the candidate set $C_t$ is too large, it will be time-consuming to calculate the probability of all actions in $C_t$, so we adopt an approximate pruning strategy. Specifically, We first select some relations with the highest probability using $p(r | s_t)$, and then select entities with the highest probability for these relations using $p(e | r, s_t)$.

For the current state $s_t$, we calculate the attention value over all relations as $p(r | s_t)$,
\begin{equation}
  \mathbf{w} = \text{Softmax}(\text{MLP}(\mathbf{s}_t) \cdot [\mathbf{r}_1, \cdots, \mathbf{r}_{|\mathcal{R}|}]) .
\end{equation}
We define a parameter $\alpha$ to control the proportion of actions that need to be added. Besides, we also have a parameter $M$ which represents the maximum number of additional actions. Therefore, the number of additional actions can be defined as,
\begin{equation}
  N_{add} = min(\lceil \alpha N \rceil, M) ,
\end{equation}
where $N$ is the action space size of the current state.
After we have the attention vector $\mathbf{w}$, we select top $x$ relations with the largest attention values in $\mathbf{w}$ to form a new relation set $\mathcal{R}_{add} = \{r^1, r^2, \cdots, r^x\}$. For every relation $r^i \in \mathcal{R}_{add}$ and the current entity $e_t$, we use the pre-trained KGE models to predict the probability distribution of the tail entity for triple query $(e_t, r^i, ?)$ as $p(e | r^i, s_t)$. We only keep the $k$ entities with the highest probability, which form $k$ additional actions $\{(r^i, e_{r^i}^{1}), \cdots, (r^i, e_{r^i}^{k})\}$ for triple query $(e_t, r^i, ?)$. Finally, all additional actions make up the additional action space $\mathcal{A}_t^{add}$ for $s_t$. Here, $k$ is a parameter, and $x$ can be calculated using previous parameters, 
\begin{equation}
  x = \lceil N_{add} / k \rceil .
\end{equation}

During the multi-hop reasoning process, we dynamically generate the additional action space $\mathcal{A}_t^{add}$ for every state $s_t$. This additional action space will be added to the original action space $\mathcal{A}_t$ and make up a new larger action space,
\begin{equation}
  \mathcal{A}_t = \mathcal{A}_t + \mathcal{A}_t^{add} .
\end{equation}
Based on the previous dynamic anticipation strategy, the dynamic completion strategy can generate more accurate action space since the state contains more prediction information. 

\subsection{Policy Optimization}

We use the typical REINFORCE  \cite{RL} algorithm to train our agent and optimize the parameters of the policy network. Specifically, the training process is obtained by maximizing the expected reward for every triple query in the training set,
\begin{equation}
  \resizebox{.85\hsize}{!}{
  $J(\theta) = \mathbb{E}_{(e_s, r, e_o) \in \mathcal{KG} } \mathbb{E}_{a_1, \dots, a_{T-1} \in \pi_{\theta} } [R(s_T|e_s, r)].$
  }
\end{equation}
The parameters $\theta$ of the policy network are optimized as follow,
\begin{equation}
  \scalebox{0.95}{$
  \begin{split}
    \nabla_{\theta}{J(\theta)} \approx \nabla_{\theta} &  {\sum_tR(s_T|e_s, r)\log{\pi_{\theta}(a_t|s_t)}} \\
    \theta &= \theta + \beta \nabla_{\theta}{J(\theta)}  ,
  \end{split}
  $}
  \label{equ:update}
  \end{equation}
where $\beta$ is the learning rate.

\begin{table}[t]
  \small
  \centering
  \scalebox{0.84}{
  \begin{tabular}{lrrrrr}
    \toprule
    \tworow{\textbf{Dataset}} & \tworow{\textbf{\#Ent}} & \tworow{\textbf{\#Rel}} & \tworow{\textbf{\#Fact}} & \twocol{\textbf{\#degree}} \\
    \cline{5-6}
      & & & & \textbf{mean} & \textbf{median} \\
      \midrule 
      FB15K-237-10\% & 11,512 & 237 & 60,985 & 5.8 & 4 \\
      FB15K-237-20\% & 13,166 & 237 & 91,162 & 7.5 & 5 \\
      FB15K-237-50\% & 14,149 & 237 & 173,830 & 13.0 & 13 \\
      NELL23K & 22,925 & 200 & 35,358 & 2.21 & 1 \\
      WD-singer & 10,282 & 135 & 20,508 & 2.35 & 2 \\
      \bottomrule
  \end{tabular}
  }
  \caption{Statistics of five datasets in experiments.}
  \label{exp:dataset_statistic}
\end{table}

\section{Experiments}

\begin{table*}[tb]
  \centering
\setlength\tabcolsep{5.5pt}
  \scalebox{0.77}{
  \begin{tabular}{lcccccccccccccccccc}
    \toprule
  \multirow{2}{*} {Model}     & \multicolumn{3}{c}{\textbf{FB15K-237-10\%}} & \multicolumn{3}{c}{\textbf{FB15K-237-20\%}} & \multicolumn{3}{c}{\textbf{FB15K-237-50\%}} & \multicolumn{3}{c}{\textbf{NELL23K}} & \multicolumn{3}{c}{\textbf{WD-singer}} \\ 
  \cmidrule(r){2-4}\cmidrule(lr){5-7}\cmidrule(lr){8-10}\cmidrule(lr){11-13}\cmidrule(lr){14-16}\cmidrule(l){17-19}
    & MRR &    @3        & @10  &MRR  &    @3        & @10  &MRR   &     @3        & @10    &MRR   &     @3        & @10       &MRR   &     @3        & @10  \\ 
  \midrule
  TransE & 10.5 & 15.9  & 27.9 & 12.3& 18.0& 31.3  & 17.7& 23.4& 40.4  & 8.4  & 10.9 & 24.7& 21.0  &32.1 & 44.6  \\
  DisMult & 7.4 & 7.5  & 16.9 & 11.3 & 11.9 & 24.0  & 18.0 & 20.2& 38.1 & 11.6 & 11.9 & 23.2& 24.4 & 27.0& 39.8  \\
  ConvE & 24.5 & 26.2 & 39.1 & 26.1 & 28.3 & 41.8 & 31.3 & \underline{34.2} & \underline{50.1} & \underline{27.6} & \underline{30.1} & 46.4 & \underline{44.8}  & \underline{47.8}& 56.9 \\
  TuckER & \underline{25.2} & \underline{27.2}& \underline{40.4}&\underline{26.6} &\underline{28.8} &\underline{42.8} & \underline{31.4} & \underline{34.2} & \underline{50.1} & 26.4& 28.9&\underline{46.7}& 42.1  &47.1& \underline{57.1} \\
  \midrule
  NeuralLP      & 7.9 & 7.2  & 13.8 & 11.2& 11.2& 17.9  & 18.2& 19.2& 24.6 & 12.2 & 13.1 & 26.3& 31.9&33.4 & 48.2   \\
  NTP      & 8.3 & 11.4  & 16.9 & 17.3& 16.1& 21.7  & 22.2& 23.1& 30.7  & 13.2 & 14.9 & 24.1& 29.2& 31.1& 44.2 \\
  MINERVA &  7.8& 7.8 & 12.2 & 15.9 & 16.4 & 22.7 & 23.0  &24.0 & 31.1 & 15.0   &15.2  &25.4  & 33.5  &37.4 & 44.9   \\ 
  MultiHopKG & 13.6 & 14.6 & 21.6 & 23.0 & 25.2  & 35.5 & 29.2 & 31.7 & 44.9 & 17.8 & 18.8 & 29.7& 35.6  & 41.1& 47.5 \\
  CPL  & 11.1 & 12.2 & 16.8 & 17.5 & 18.4 & 25.7 & 26.4 & 28.5 & 36.8 & -  &-  &-  & 34.2    &40.1 & 46.3 \\
  \midrule
  DacKGR (sample) & 21.8 & \textbf{23.9} & \textbf{33.7} & \textbf{24.7} & \textbf{27.2}  & \textbf{39.1} & \textbf{29.3} & \textbf{32.0} & 45.7 & \textbf{20.1}  & \textbf{21.6} & \textbf{33.2} & \textbf{38.1}  & \textbf{42.3} & \textbf{50.6}  \\ 
  DacKGR (top) & \textbf{21.9} & \textbf{23.9} & 33.5 & 24.4 & 27.1  & 38.9 & \textbf{29.3} & 31.8 & \textbf{45.8} & 19.1  & 20.0 & 30.8 & 37.0  & 40.5 & 46.5 \\ 
  DacKGR (avg) & 21.5 & 23.2 & 33.4 & 24.2 & 26.6  & 38.8 & 29.1 & 31.9 & 45.4 & 17.1  & 18.6 & 28.2 & 36.4  & 40.1 & 48.0  \\ 
  \bottomrule
  \end{tabular}
  }
  \vspace{-0.1cm}
  \caption{Link prediction results on five datasets from Freebase, NELL and Wikidata. @3 and @10 denote Hits@3 and Hits@10 metrics, respectively. All metrics are multiplied by 100. The best score of multi-hop reasoning models is in \textbf{bold}, and the best score of embedding-based models is \underline{underlined}.}
  \label{fb15k-lp}
  \vspace{-0.2cm}
  \end{table*}

\subsection{Datasets}

In this paper, we use five datasets sampled from Freebase \cite{Freebase}, NELL \cite{NELL} and Wikidata \cite{Wikidata} for experiments. Specifically, in order to study the performance of our method on KGs with different degrees of sparsity, we constructed three datasets based on FB15K-237 \cite{FB15K-237}, i.e., FB15K-237-10\%, FB15K-237-20\% and FB15K-237-50\%. These three datasets randomly retain 10\%, 20\% and 50\% triples of FB15K-237 respectively. 

In addition, we also construct two datasets NELL23K and WD-singer from NELL and Wikidata, where WD-singer is a dataset of singer domain from Wikidata. For NELL23K, we first randomly sample some entities from NELL and then sample triples containing these entities to form the dataset. For WD-singer, we first find all concepts related to singer in Wikidata, then use the entities corresponding to these concepts to build the entity list. After that, we expand the entity list appropriately, and finally use the triples containing entities in the entity list to form the final dataset. The statistics of our five datasets are listed in Table \ref{exp:dataset_statistic}.

\subsection{Experiment Setup}

\textbf{Baseline Models} \indent In our experiments, we select some KGE models and multi-hop reasoning models for comparison. For embedding-based models, we compared with TransE \cite{TransE}, DistMult \cite{DistMult}, ConvE \cite{ConvE} and TuckER \cite{TuckER}. For multi-hop reasoning, we evaluate the following five models \footnote{M-walk does not provide the necessary source codes and we do not compare with it.},  Neural Logical Programming (NeuralLP) \cite{NeuralLP},  Neural Theorem Prover (NTP) \cite{NTP}, MINERVA \cite{MINERVA}, MultiHopKG \cite{MultiHop} and CPL \footnote{CPL can not run on NELL23K since its entities do not have additional text information.} \cite{CPL} .
Besides, our model has three variations, DacKGR (sample), DacKGR (top) and DacKGR (avg), which use sample, top-one and average strategy (introduced in Section \ref{sec:state_prediction}) respectively.

\noindent \textbf{Evaluation Protocol} \indent For every triple $(e_s, r_q, e_o)$ in the test set, we convert it to a triple query $(e_s, r_q, ?)$, and then use embedding-based models or multi-hop reasoning models to get the ranking list of the tail entity. Following the previous work \cite{TransE}, we use the ``filter" strategy in our experiments.
We use two metrics: (1) the mean reciprocal rank of all correct tail entities (MRR), and (2) the proportion of correct tail entities ranking in the top K (Hits@K) for evaluation.

\noindent \textbf{Implementation Details} \indent In our implementation, we set the dimension of the entity and relation vectors to 200, and use the ConvE model as the pre-trained KGE for both dynamic anticipation and dynamic completion strategies. In addition, we use a 3-layer LSTM and set its hidden dimension to 200. Following previous work \cite{MINERVA}, we use Adam \cite{Adam} as the optimizer. For the parameters $\alpha, M$ and $k$ in the dynamic completion strategy, we choose them from \{0.5, 0.33, 0.25, 0.2\}, \{10, 20, 40, 60\} and \{1, 2, 3, 5\} respectively. We select the best hyperparameters via grid search according to Hits@10 on the valid dataset. Besides, for every triple $(e_s, r_q, e_o)$ in the training set, we also add a reverse triple $(e_o, r_q^{inv}, e_s)$.

\subsection{Link Prediction Results}

The left part of Table \ref{fb15k-lp} shows the link prediction results on FB15K-237-10\%, FB15K-237-20\% and FB15K-237-50\%. From the table, we can learn that our model outperforms previous multi-hop reasoning models on these three datasets, especially on FB15K-237-10\%, where our model gains significant improvements compared with the best multi-hop reasoning baseline MultiHopKG (which is about 56.0\% relative improvement on Hits@10). 

When we compare the experimental results on these three datasets horizontally (from right to left in Table \ref{fb15k-lp}), we can find that as the KG becomes sparser, the relative improvement of our model compared with the baseline models is more prominent. This phenomenon shows that our model is more robust to the sparsity of the KG compared to the previous multi-hop reasoning model. 

As shown in previous work \cite{MultiHop, CPL}, KGE models often achieve better results than multi-hop reasoning models. This phenomenon is more evident on sparse KGs. The results of these embedding-based models are only used as reference because they are different types of models from multi-hop reasoning and are not interpretable.

The right part of Table \ref{fb15k-lp} shows the link prediction results on NELL23K and WD-singer. From the table, we can find a phenomenon similar to that in the left part of Table \ref{fb15k-lp}. Our model performs better than previous multi-hop reasoning models, which indicates that our model can be adapted to many other knowledge graphs.

From the last three rows of Table \ref{fb15k-lp}, we can learn that the sample strategy in Section \ref{sec:state_prediction} performs better than top-one and average strategies in most cases. This is because these two strategies lose some information. The top-one strategy only retains the entity with the highest probability. The average strategy uses a weighted average of entity vectors, which may cause the features of different vectors to be canceled out.

\subsection{Ablation Study}

\begin{table}[h]
  \centering
  \scalebox{0.75}{
    \begin{tabular}{ccccc}
      \toprule[1pt]
    Model& MRR & Hits@1 & Hits@3 & Hits@10 \\
    \midrule[0.5pt]
    MultiHopKG & 23.0 & 16.9 & 25.2 & 35.5 \\
    DacKGR (w/o DC) & 23.5 & 17.2 & 25.8 & 36.5 \\
    DacKGR (w/o DA) &  24.2 & 17.7 & 26.6 & 37.9 \\
    DacKGR & \textbf{24.7} & \textbf{17.8} & \textbf{27.2} & \textbf{39.1} \\
    \bottomrule[1pt]
    \end{tabular}
    }
    \caption{Ablation study results on FB15K-237-20\%. DC and DA denote dynamic completion and dynamic anticipation strategy respectively.}
      \label{ablation_study}
  \end{table}

  \begin{figure*}[ht]
    \centering
    \setlength{\abovecaptionskip}{2pt}
    \setlength{\belowcaptionskip}{0pt}
    \includegraphics[width=155mm]{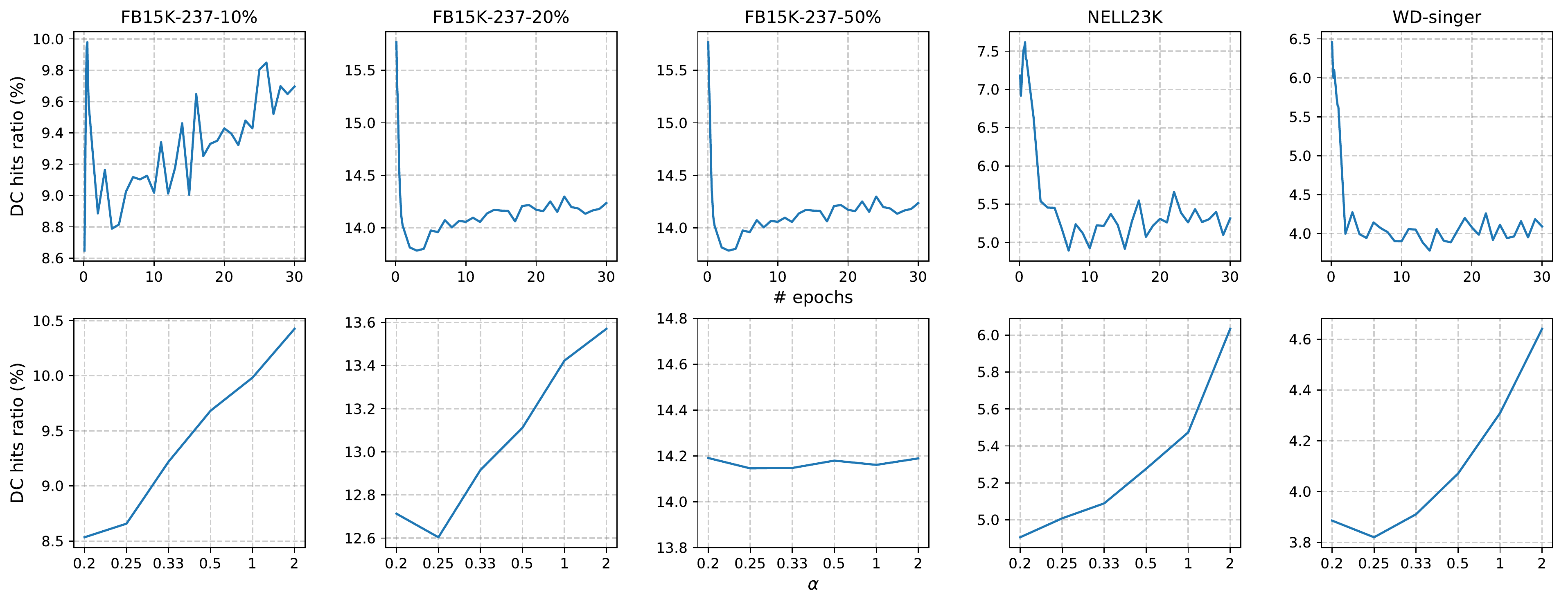}
    \caption{The top and bottom rows show the DC hits ratio change w.r.t. \#epoch and $\alpha$ respectively.} 
    \label{pic:parameters} 
    \end{figure*}

    \begin{table*}[ht]
    \centering
    \scalebox{0.91}{
     \begin{tabular}{c|c}
      \toprule[1pt]
      \multicolumn{2}{c}{Triple query: \textit{(Kirkby Lunn, country of citizenship, ?)}} \\
      \midrule
      1 & \textit{Kirkby Lunn} $\xrightarrow{\textit{\textbf{place of burial}}}$ \textit{\textbf{London}} $\xrightarrow{\textit{country}}$ \textit{United Kingdom} $\xrightarrow{\textit{LOOP}}$ \textit{\underline{United Kingdom}} \\
      2 & \textit{Kirkby Lunn} $\xrightarrow{\textit{student of}}$ \textit{Jacques Bouhy} $\xrightarrow{\textit{country of citizenship}}$ \textit{Belgium} $\xrightarrow{\textit{LOOP}}$ \textit{Belgium} \\
      3 & \textit{Kirkby Lunn} $\xrightarrow{\textit{student of}}$ \textit{Albert Visetti} $\xrightarrow{\textit{student}}$ \textit{Agnes Nicholls} $\xrightarrow{\textit{country of citizenship}}$ \textit{\underline{United Kingdom}} \\
      \bottomrule[1pt]
      \end{tabular}
    }
      \caption{Case study of our model on link prediction experiment. For the triple query, we show the three reasoning paths with the top-3 scores via beam search. The relation and entity in \textbf{bold} are additional actions generated using dynamic completion strategy. The correct entities for the triple query are \underline{underline}. }
    \label{table:case_study}
   \end{table*}

In this paper, we design two strategies for sparse KGs. In order to study the contributions of these two strategies to the performance of our model, we conduct an ablation experiment by removing dynamic anticipation (DA) or dynamic completion (DC) strategy on FB15K-237-20\% dataset. 

As shown in Table \ref{ablation_study}, removing either the DA or DC strategy will reduce the effectiveness of the model, which demonstrates that both strategies contribute to our model. Moreover, we can learn that using either strategy individually will enable our model to achieve better results than the baseline model. Specifically, the model using the DC strategy alone performs better than the model using the DA strategy alone, which is predictable, since the DA strategy only allows the agent to make a correct choice, and will not substantially alleviate the sparsity of KGs.

\subsection{Analysis}

In the dynamic completion (DC) strategy, we dynamically provide some additional actions for every state, which enrich the selection space of the agent and ease the sparsity of KGs. However, will the agent choose these additional actions, or in other words, do these additional actions really work? 

In this section, we analyze the results of the DC hits ratio, which indicates the proportion of the agent selecting additional actions (e.g., choosing actions in $\mathcal{A}_t^{add}$ for $s_t$). In the first step, we analyze the change of DC hits ratio as the training progresses, which is shown in the first row of Figure \ref{pic:parameters}. From this figure, we can learn that for most KGs (except FB15K-237-10\%), DC hits ratio is relatively high at the beginning of training, then it will drop sharply and tend to stabilize. This is reasonable because there is some noise in the additional actions. In the beginning, the agent cannot distinguish the noise part and choose them as the same as the original action. But as the training proceeds, the agent can identify the noise part, and gradually reduces the selection ratio of additional actions. For FB15K-237-10\%, DC hits ratio will decrease first and then increase. This is because many triples have been removed in FB15K-237-10\%, which exacerbates the incompleteness of the dataset. The additional actions work more effectively in this situation and increase the probability of correct reasoning. 

In the second row of the Figure \ref{pic:parameters}, we give the effect of parameter $\alpha$ (indicates the proportion of actions that need to be added) described in Section \ref{sec:action_selection} on the DC hits ratio. Specifically, We use the average DC hits ratio results of the last five epochs as the final result. From this figure, we can find that for most datasets, DC hits ratio will gradually increase as $\alpha$ increases. This is as expected because a larger $\alpha$ means more additional actions, and the probability that they are selected will also increase. It is worth noting that on the FB15K-237-50\%, DC hits ratio hardly changes with $\alpha$. This is because the sparsity of FB15K-237-50\% is not severe and does not rely on additional actions.

\subsection{Case Study}

In Table \ref{table:case_study}, we give an example of triple query and three reasoning paths with the top-3 scores given by our model DacKGR. From the first path, we can learn that our dynamic completion strategy can provide agents with some additional actions that are not in the dataset, and further form a reasoning path. Besides, as shown in the third path, DacKGR can also use the paths that exist in the KG to perform multi-hop reasoning.

\section{Related Work}

\subsection{Knowledge Graph Embedding}

Knowledge graph embedding (KGE) aims to represent entities and relations in KGs with their corresponding low-dimensional embeddings. It then defines a score function $f(e_s, r_q, e_t)$ with embeddings to measure the correct probability of each triple. Specifically, most KGE models can be divided into three categories \cite{wang2017knowledge}: (1) Translation-based models \cite{TransE, TransH, TransR, RotatE} formalize the relation as a translation from a head entity to a tail entity, and often use distance-based score functions derived from these translation operations. (2) Tensor-factorization based models \cite{RESCAL, DistMult, TuckER} formulate KGE as a three-way tensor decomposition task and define the score function according to the decomposition operations. (3) Neural network models \cite{NTN, ConvE, ConvKB, R-GCN} usually design neural network modules to enhance the expressive abilities. Generally, given a triple query $(e_s, r_q, ?)$, KGE models select the entity $e_o$, whose score function $f(e_s, r_q, e_o)$ has the highest score as the final prediction. Although KGE models are efficient, they lack interpretability of their predictions.

\subsection{Multi-Hop Reasoning}

Different from embedding-based models, multi-hop reasoning for KGs aims to predict the tail entity for every triple query $(e_s, r_q, ?)$ and meanwhile provide a reasoning path to support the prediction. 
Before multi-hop reasoning task is formalized, there are some models on 
relation path reasoning task, which aims to predict the relation between entities like $(e_s, ?, e_o)$ using path information. DeepPath \cite{DeepPath} first adopts reinforcement learning (RL) framework for relation path reasoning, which inspires much later work (e.g., DIVA \cite{DIVA} and AttnPath \cite{ATTPath}). 

MINERVA \cite{MINERVA} is the first model that uses REINFORCE algorithm to do the multi-hop reasoning task. To make the training process of RL models stable, \citeauthor{M-Walk} propose M-Walk to solve the reward sparsity problem using off-policy learning. MultiHopKG \cite{MultiHop} further improves MINERVA using action dropout and reward shaping. \citet{MetaKGR} propose MetaKGR to address the new task that multi-hop reasoning on few-shot relations. In order to adapt RL models to a dynamically growing KG, \citet{CPL} propose CPL to do multi-hop reasoning and fact extraction jointly. In addition to the above RL-based reasoning models, there are some other neural symbolic models for multi-hop reasoning. NTP \cite{NTP} and NeuralLP \cite{NeuralLP} are two end-to-end reasoning models that can learn logic rules from KGs automatically. 

Compared with KGE models, multi-hop reasoning models sacrifice some accuracy for interpretability, which is beneficial to fine-grained guidance for downstream tasks.

\section{Conclusion}

In this paper, we study the task that multi-hop reasoning over sparse knowledge graphs. The performance of previous multi-hop reasoning models on sparse KGs will drop significantly due to the lack of evidential paths. In order to solve this problem, we propose a reinforcement learning model named DacKGR with two strategies (i.e., dynamic anticipation and dynamic completion) designed for sparse KGs. These strategies can ease the sparsity of KGs. In experiments, we verify the effectiveness of DacKGR on five datasets. Experimental results show that our model can alleviate the sparsity of KGs and achieve better results than previous multi-hop reasoning models.
However, there is still some noise in the additional actions given by our model. In future work, we plan to improve the quality of the additional actions.

\section*{Acknowledgments}

This work is supported by NSFC Key Projects (U1736204, 61533018), a grant from Institute for Guo Qiang, Tsinghua University (2019GQB0003), Alibaba and THUNUS NExT Co-Lab.

\bibliography{emnlp2020}
\bibliographystyle{acl_natbib}

\end{document}